\documentclass[10pt,twocolumn,letterpaper]{article}

\usepackage[pagenumbers]{cvpr} 

\usepackage{float}
\usepackage{makecell}
\usepackage{multirow}
\usepackage[most]{tcolorbox}

\newtcblisting{atformat}{
	listing only,
	colback=black!2,
	colframe=black!25,
	arc=1.5mm,
	boxrule=0.4pt,
	top=2pt,bottom=2pt,left=3pt,right=3pt,
	listing options={
		basicstyle=\ttfamily\footnotesize,
		columns=fullflexible,
		breaklines=true
	}
}

\newcommand{\oursicon}{ART}
\newcommand{\oursdata}{AT}
\usepackage{algorithm}
\usepackage{algpseudocode}

\definecolor{cvprblue}{rgb}{0.21,0.49,0.74}
\usepackage[pagebackref,breaklinks,colorlinks,allcolors=cvprblue]{hyperref}

\def\paperID{N/A} 
\def\confName{CVPR}
\def\confYear{2026}

\title{Evolve Vision-Language-Action Model into an Agent with On-the-fly Tool-use}

\author{
	Yi Ding$^{1}$\thanks{Equal contribution. Work done during internship at Astribot.} \quad
	Yanzhao Yu$^{1,2}$\footnotemark[1] \quad
	Xili Dai$^{3}$ \quad
	Xianbiao Qi$^{4}$ \quad \\
	Peiwen Sun$^{5}$ \quad
	Xueqian Wang$^{2}$ \quad
	Xiangyu Yue$^{5}$ \quad
	Jianan Wang$^{1}$\thanks{Corresponding author.} \\
	$^{1}$ Astribot \quad
	$^{2}$ Tsinghua University \quad
	$^{3}$ Juxi Tech \\
	$^{4}$ IntelliFusion Inc. \quad
	$^{5}$ The Chinese University of Hong Kong
	\\
	{\tt\small onedonedone@sjtu.edu.cn, yuyz24@mails.tsinghua.edu.cn, jiananwang@astribot.com}
}

\begin{document}
\maketitle

\begin{figure*}[htbp]
	\centering
	\includegraphics[width=\textwidth]{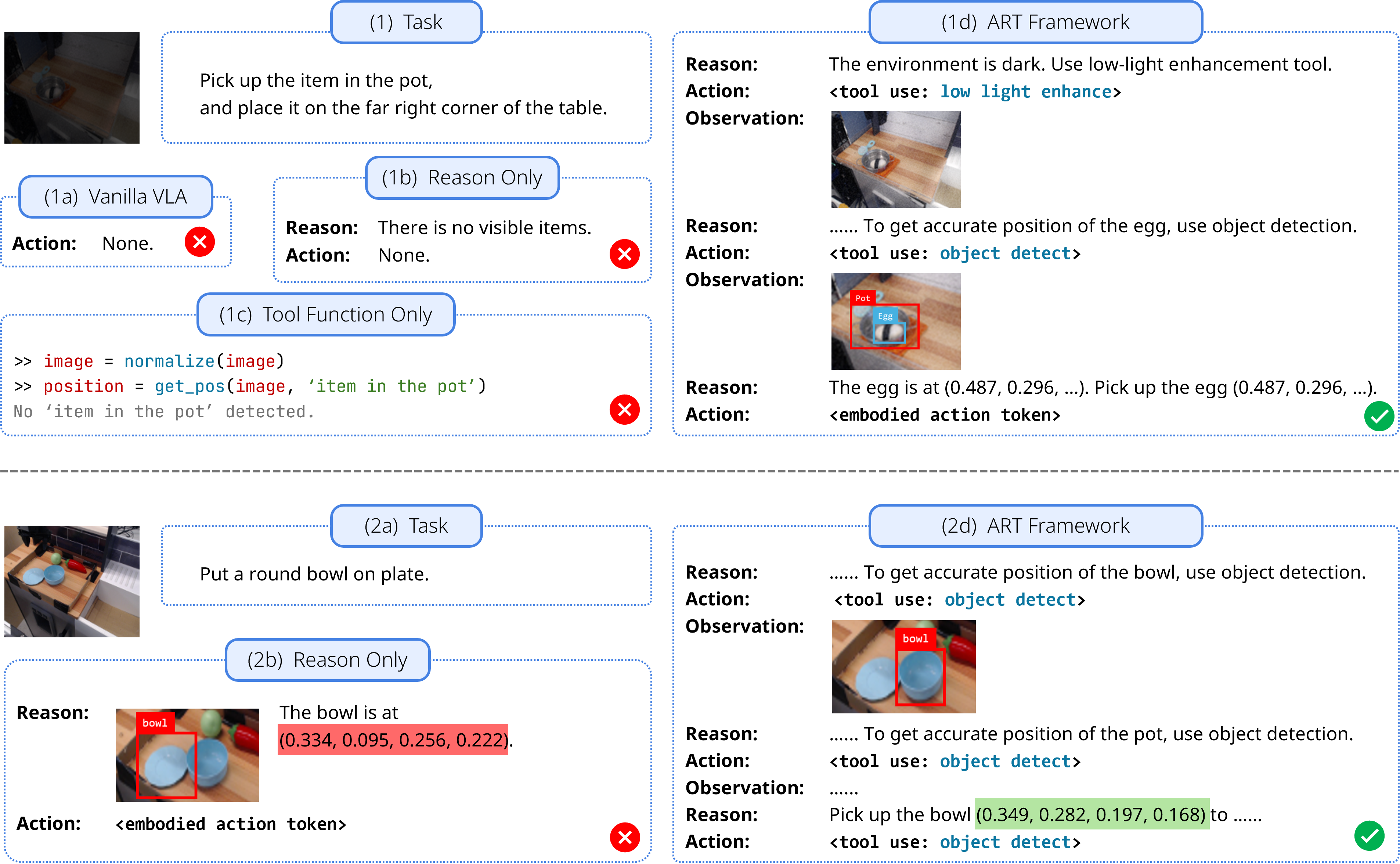}
	\caption{\textbf{Comparison of VLA Paradigms}, including (a) standard end-to-end VLA \cite{OpenVLA, PI-0, RT-2}, (b) VLA enhanced with Chain-of-Thought reasoning \cite{ECoT}, (c) modular robotic behavior synthesis \cite{CaP, RoboCodeX}, and (d) our Agentic Robot with Tool use (\oursicon{}) framework, which intDegrates tool use into the end-to-end action generation process. The \oursicon{} model demonstrates enhanced adaptability to complex environments (Task 1) and correct hallucinated predictions in foundation models (Task 2) through dynamic tool use.}
	\label{Fig: Pipeline Comparison.}
\end{figure*}

\begin{abstract}
	This paper integrates end-to-end Visual-Language-Action (VLA) models with agentic tool-use to propose \textbf{A}gentic \textbf{R}obot with \textbf{T}ool-use (ART). ART is a tool-injection framework that tunes any VLA model to leverage off-the-shelf tool modules for low-level vision, high-level affordance, and embodiment enhancement. Compared to vanilla VLA models with a whole continuous action solution space, ART reduces the complexity of the action solution space through tool-use, which not only improves generalizability across different tasks but also reduces data dependency. To demonstrate the advantages (high generalizability and low data dependency) of this framework, we first built a dataset of 30K tool-use trajectories and action demonstrations, which is much smaller than those used by baseline methods. We then designed a training regimen for long-trajectory tool-use reasoning in challenging environments. Experiments show that ART achieves a 20\% higher success rate than mainstream baselines on simulation and real-world tasks, such as pick-and-place in the dark at novel viewpoints. Empirical results highlight the benefits of an agent-based approach: modular tool utilization enables more efficient training, lightweight deployment, and scalable integration of new tools. This design fosters robustness, adaptability, and extensibility, paving the way for the practical deployment of VLA systems in complex real-world scenarios.
\end{abstract}

\section{Introduction}

The field of Vision-Language-Action (VLA) models has seen significant advancements, broadly categorized into two main development paths: modular and end-to-end approaches. The \textbf{modular approach}, exemplified by methods such as RoboTool \cite{RoboTool}, RoboScript \cite{RoboScript}, and RoboCodeX \cite{RoboCodeX}, encapsulates specific functionalities like action execution and goal recognition within distinct tool modules. Models are then trained to call the APIs of these predefined tools. While this approach offers complete decoupling of different capabilities, its reliance on fixed tool sets and action functions inherently limits its applicability, primarily to simpler action outputs. In contrast, the \textbf{end-to-end approach}, as demonstrated by models like OpenVLA \cite{OpenVLA}, ECoT \cite{ECoT}, and \(\pi\) \cite{PI-0, PI-0.5, PI-0.5-KI, PI-FAST}, involves training a single base model on extensive multi-task datasets. This paradigm allows models to produce highly precise and detailed actions in these trained tasks. However, a significant drawback arises when encountering novel scenarios and tasks: the base model typically necessitates complex and costly post-training. Furthermore, the inherent coupling of all capabilities in end-to-end models makes them susceptible to catastrophic forgetting when adapted to new data.

\noindent A critical question thus emerges: \textit{can we harness the powerful, precise action output capabilities of end-to-end VLA models while simultaneously enabling them to swiftly integrate and utilize external tools for adapting to new scenarios and tasks?}

Naturally, the action solution space of a VLA model would be significantly constrained if continuous actions were replaced by predefined discrete tools. Hence, in this paper, we propose \textbf{ART (Agentic Robot with Tool-use)}, a novel fine-tuning tool-injection framework for VLA models that seamlessly integrates multi-modal tool usage. Through ART, VLA models can acquire new, plug-and-play capabilities—such as visual enhancement, sophisticated spatial reasoning, and expanded action abilities—without compromising their original action output fidelity. Compared to vanilla VLA models with a whole continuous action solution space, ART framework reduces the complexity of the action solution space through tool-use, which not only improves generalizability across different tasks but also reduces data dependency. 

To demonstrate the advantages (high generalizability and low data dependency) of this framework, we first built a dataset of 30K tool-use trajectories and action demonstrations, which is much smaller than those used by baseline methods. Inspired by T3-Agent \cite{T3-Agent}, we introduce a novel pipeline to extend existing VLA datasets. This three-step process—task design, reasoning generation, and tool-trajectory synthesis—efficiently generates large volumes of tool-usage data. Initially, we systematically "degrade" existing VLA data by introducing complexities into task instructions, thereby creating tasks that explicitly necessitate tool usage for their resolution. Subsequently, based on these modified tasks, we leverage the base VLA model to generate the requisite reasoning processes for invoking necessary tools. Finally, these reasoning processes are synthesized into comprehensive tool usage trajectories. This innovative approach allows for the integration of tool reasoning into existing datasets without the need for costly manual collection of new action data. Then, to mitigate the risk of catastrophic forgetting during fine-tuning with new tool-use data, we define a two-stage LoRA fine-tuning method. During the tool usage training phase, the original VLA model weights are frozen, and only the LoRA modules dedicated to tool reasoning are fine-tuned. Crucially, once the VLA model has utilized the tools, we mask out the LoRA layer's outputs during action generation, allowing the model to revert to its original action output capabilities. This strategic decoupling effectively separates tool reasoning from the core VLA model, thereby preventing data conflicts and preserving high-quality action generation during training.

In summary, our key contributions are as follows:
\begin{itemize}
	\item We introduce a data generation method that integrates multi-modal tools into existing VLA datasets, enabling complex tool usage in long task chains.
	\item We propose ART (Agentic Robot with Tool-use), a fine-tuning tool-injection framework for VLA models that preserves the pre-trained VLA action generation ability while enabling new tool reasoning capabilities.
	\item Through experiments in the LIBERO simulation and real-world settings, we demonstrate that our method enhances the model's robustness in handling new scenarios, objects, and action instructions.
\end{itemize}

\section{Related Works}

We first provide an overview of two research directions in VLA models, followed by a brief discussion of tool use in multi-modal agents, another cornerstone of \oursicon{}.

\paragraph{Modular Robotic Behavior Synthesis.} The fundamental challenge in early-stage research lies in bridging the gap between high-level human instructions and robotic behaviors. A mainstream research line breaks down action generation into modular subtasks \cite{CaP,Instruct2Act,ProgPrompt,RoboCodeX,RoboOS,RoboScript,RoboTool,Text2Motion}. Given the observation \(\boldsymbol{o}_t\) at time \(t\), the model \(\pi\) synthesizes a sequence of external modules that progressively address the task, rather than directly generating actions:
\begin{equation} 
	\pi : \boldsymbol{o}_t \mapsto (f, f_\mathrm{act}), \quad f : \boldsymbol{o}_t \mapsto \boldsymbol{o}^*_t, \enspace f_\mathrm{act} : \boldsymbol{o}^*_t \mapsto \boldsymbol{A}_t
\end{equation}
where given the environment observations \(\boldsymbol{o}_t\), \(f\) returns intermediate representations, referred to as affordances, such as object positions \cite{MDETR, ViLD} and segmentation masks \cite{SAM}. \(f_\mathrm{act}\) denotes the embodied functions that translate the geometric position to robot movements \cite{RoboOS}. For example, CaP \cite{CaP} invokes the function \(\mathtt{pick\_obj}\) and \(\mathtt{place\_at\_pos}\) in sequence to complete a pick-and-place task. The method packs up different robotic abilities as modules, and then the model can flexibly utilize off-the-shelf tools. However, the method oversimplifies the language-to-action mapping into handcrafted functions, and it struggles with complex tasks requiring complicated language understanding and dexterous action generation \cite{SurveyPKU}.

\begin{figure*}[htbp]
	\centering
	\includegraphics[width=\textwidth]{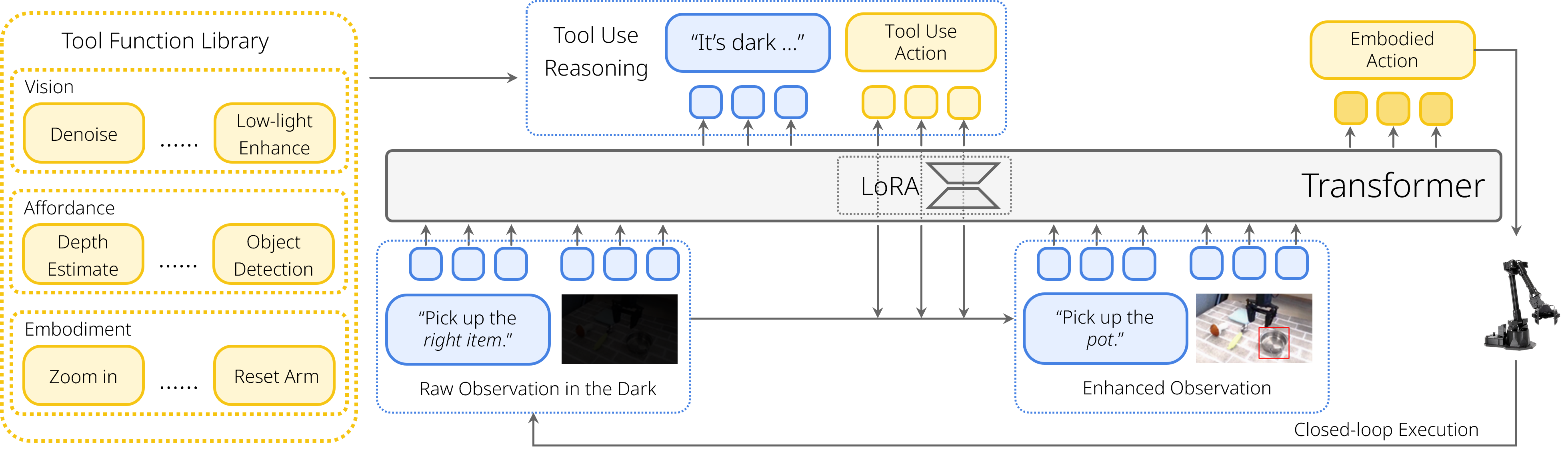}
	\caption{\textbf{Overview of the \oursicon{} architecture.} With a fine-tuned LoRA module, \oursicon{} first reasons over raw observations and predicts tool use actions that activate external tools, thereby improving the input data. Specifically, the external tools provide vision, affordance, and embodiment enhancement. Given the enhanced observations, the model generates the final embodied actions.}
	\label{Fig: Overall Framework}
\end{figure*}

\paragraph{Vision-Language-Action Model.} The success of large-scale multi-modal models \cite{Flamingo,PaLI-X,Llama-2,PaLM-E,PaliGemma} has motivated an end-to-end method to generate generalist robot policies \cite{Octo,OpenVLA,PI-0,PI-0.5,PI-0.5-KI,PI-FAST,RoboFlamingo,RT-1,RT-2}, where a foundation model \(\pi\) is finetuned to directly predict an action chunk \(\boldsymbol{A}_t\) conditioned on the observation \(\boldsymbol{o}_t\) at time \(t\), written as
\begin{equation}\label{Eq: Standard VLA}
	\pi : \boldsymbol{o}_t \mapsto \boldsymbol{A}_t, \quad \boldsymbol{A}_t= [\boldsymbol{a}_t,\boldsymbol{a}_{t+1},\cdots,\boldsymbol{a}_{t+N}],
\end{equation}
where the observation \(\boldsymbol{o}_t\) consists of the language instructions \(\boldsymbol{l}\), visual data \(\boldsymbol{I}_t\), and the proprioceptive states \(\boldsymbol{q}_t\) of the robot, and \(\boldsymbol{a}_t\) is a continuous control signal. For a robot with \(d\) degrees of freedom, \(\boldsymbol{a}_t \in \mathbb{R}^d\). Pre-trained on multi-modal web data \cite{Cambrian-1,CapsFusion,COCO,PaLI,VQAv2}, and co-trained on several robot datasets \cite{BridgeData-v2,DROID,LIBERO,Open-X}, this end-to-end model can directly map human instructions to robotic actions, and generalize to multiple scenes \cite{PI-0.5,RAD,RT-2}.

To improve planning ability and action accuracy, following methods tailor the model \(\pi\) to describe intermediate affordance \(\boldsymbol{o}^*_t\), such as object bounding boxes \cite{ECoT,ECoT-Lite} or manipulation keypoints \cite{RoboBrain}. These predictions serve as guidance for subsequent action generation, formalized as
\begin{equation}
	\pi = \pi_2 \circ \pi_1, \quad \pi_1 : \boldsymbol{o}_t \mapsto \boldsymbol{o}^*_t, \enspace \pi_2 : \left(\boldsymbol{o}_t, \boldsymbol{o}^*_t\right) \mapsto \boldsymbol{A}_t,
\end{equation}
However, the approach has several limitations. First, it relies on large-scale data with handcrafted affordance annotations that are hard to come by in robotics \cite{Open-X}. Second, it depends on high-quality observations and is vulnerable to visual perturbations which significantly degrade their performance in realistic scenarios \cite{LIBERO-Plus}.

\paragraph{Multi-modal Agents with Tool-use.} Research on large language models demonstrates their ability as multi-modal agents to solve complex tasks with tool use \cite{CLOVA, CRAFT, ReAct, SeeAct, Steve-Eye, T3-Agent, ViperGPT, VisualProgramming}. The agent leverages powerful language models to generate pseudocode \cite{VisualProgramming, CLOVA}, scripts \cite{ViperGPT}, or a API list \cite{HuggingGPT} that invokes external tools in sequence. Given the model's generalist reasoning ability and the external tools' enhancement, agents show superior performance in the web search \cite{SeeAct}, image editing \cite{VisualProgramming}, and game play \cite{Steve-Eye}. Recently, several works have incorporated tool-use into the end-to-end VLA framework. TIGeR \cite{TIGeR} incorporates geometric computation tools into end-to-end action generation to enhance the VLA's spatial perception and reasoning capabilities, and VLA\textsuperscript{2} \cite{VLAA} employs web knowledge retrieval to generalize the VLA model on tasks involving unseen targets. These methods demonstrate the scalability and flexibility of combining the end-to-end architecture with on-the-fly external tools. However, they are limited on specific tasks, object identification \cite{VLAA} and 3D geometrics \cite{TIGeR}. In contrast, \oursicon{} takes on the challenge of tool use in long, multi-task trajectories, thereby facilitating the solution of a broader range of tasks with tool-integrated reasoning.

\section{ART: Agentic Robot with Tool-use}

In this section, we present our framework for an agentic robot with tool use. We begin with the formulation of optimization objective (Sec.~\ref{Sec: Optimization with Tool-use Reasoning}). Next, we detail the system architecture designed for tool injection (Sec.~\ref{Sec: Tool Injection Framework}), followed by a brief description of our training procedures (Sec.~\ref{Sec: Training Procedures}).

\subsection{Optimization with Tool-use Reasoning}\label{Sec: Optimization with Tool-use Reasoning}

We begin with the standard setup of a vanilla VLA model. At each timestep \(t\), the model \(\pi\) receives a clear and high-quality observation \(\boldsymbol{o}_t\) as input, from which it predicts an embodied action \(\boldsymbol{a}_t \in \mathcal{A}\) that governs the robot's movement. However, in real-world scenarios, observations are often ambiguous and prone to various forms of corruption, which impedes the model to accurately learn the mapping from observations to embodied actions during training.

The idea behind \oursicon{} is simple: we augment the model's action space to \(\mathcal{A}^* = \mathcal{A} \times \mathcal{R} \times \mathcal{T}\), where \(\mathcal{R}\) represents the space for language reasoning, and \(\mathcal{T}\) is the action space for tool usage. An action \(\boldsymbol{a}_{\mathrm{r},t} \in \mathcal{R}\) aims to compose information by reasoning over the observations, while \(\boldsymbol{a}_{\mathrm{t},t} \in \mathcal{T}\) is intended to activate external tools to enhance the observations. Thus, a solution to a task is represented as a trajectory of augmented actions, denoted as \(\boldsymbol{a}^*_{1:T} = \left(\boldsymbol{a}^*_1, \boldsymbol{a}^*_2, \dots, \boldsymbol{a}^*_T\right)\), where each \(\boldsymbol{a}^*_t \in \mathcal{A}^*\). Consequently, the overall optimization objective becomes:
\begin{equation}
	\pi^* = \mathop{\arg\max}\limits_{\pi} \, P_\theta \left( \boldsymbol{a}^*_{1:T} \mid \tilde{\boldsymbol{o}}_{1:T} \right),
\end{equation}
where the model \(\pi\) is parameterized by \(\theta\).

Our primary insight is that an accurate embodied action \(\boldsymbol{a}_t\) should be independent of the noise and corruptions in raw observations \(\tilde{\boldsymbol{o}}_t\), and thus depend solely on the enhanced observations \(\boldsymbol{o}_t\). Thus, the overall optimization objective is decomposed as
\begin{equation}\label{Eq: Optimization Objective Decomposition}
	\pi^* = \mathop{\arg\max}\limits_{\pi} \prod_{t=1}^T P_{\theta,1}(\boldsymbol{a}_t | \boldsymbol{o}_{1:t}) P_{\theta,2}(\boldsymbol{a}_{\mathrm{r},t}, \boldsymbol{a}_{\mathrm{t},t} | \tilde{\boldsymbol{o}}_{1:t}).
\end{equation}

Therefore, we decompose the overall objective into two: \(P_{\theta,1}\), the training objective of a vanilla VLA, and \(P_{\theta,2}\), the fine-tuning objective of reasoning and tool use, which we refer to as tool injection.

\subsection{Tool Injection Framework}\label{Sec: Tool Injection Framework}

The decomposition in Sec.~\ref{Sec: Optimization with Tool-use Reasoning} allows us to train \oursicon{} from fine-tuning an existing VLA model with an extra optimization objective. However, previous works prove that a naïve coupling of training recipes significantly degrades the capability of the backbone model and causes catastrophic forgetting \cite{PI-0.5-KI,SurveyPKU}. To avoid it and realize the factorization in Eq.~\ref{Eq: Optimization Objective Decomposition}, we introduce a \emph{non-destructive} modification within the standard VLA architecture.

\paragraph{Tool Token Injection.} Unlike an embodied action \(\boldsymbol{a}_t\) that encodes a \emph{continuous} control signal, the tool use action \(\boldsymbol{a}_{\mathrm{t},t}\) is \emph{discrete}, as the state of each tool is naturally binary: either active or inactive. Given a tool function set \(\mathcal{F}\) of size \(n\), an action \(\boldsymbol{a}_{\mathrm{t},t} \in \{0,1\}^n\), where each element \(a_{\mathrm{t},t,i}\) indicates the state of the \(i\)-th tool function \(f_i \in \mathcal{F}\). This discrete representation enables us to map tool use actions to tokens in the vocabulary, allowing the model to treat tool activation as a form of discrete decision-making. By representing the tool states as tokens, we treat tool use similarly to language generation, where each token corresponds to the activation or deactivation of a specific tool function. To minimize the impact on the original vocabulary, following RT-2 \cite{RT-2} and OpenVLA \cite{OpenVLA}, we leverage the last \(N\) tokens in the vocabulary for this purpose. Consequently, tool use tokens are seamlessly integrated into the standard next-token prediction training, with the cross-entropy loss calculated on the predicted actions.

\paragraph{Adaptive LoRA Fine-tuning.} To prevent interference between embodied action generation and tool use, we adopt a fine-tuning strategy with a dynamically activated LoRA module \cite{LoRA}. Starting from a pre-trained VLA model, we freeze the backbone and fine-tune the LoRA module to enhance tool-use reasoning. This decouples the two components: the VLA backbone without LoRA generates embodied actions, while with the LoRA module, the model generates tool use actions. In the inference process, the model first generates reasoning and tool use actions (\(\boldsymbol{a}_{\mathrm{r},t}\), \(\boldsymbol{a}_{\mathrm{t},t}\)) based on observations, activating tools with LoRA; then, the enhanced observation \(\boldsymbol{o}_t\) is passed to the model to generate the final embodied action \(\boldsymbol{a}_t\). This independent optimization of tool use and action generation ensures efficient learning and allows ART to be easily integrated into various VLA architectures with minimal modifications.

\paragraph{Tool‑use with Action Chunk.} Modern VLA architectures increasingly adopt an action‑chunking strategy to improve efficiency, where the model predicts a chunk of \(H\) future embodied actions in one inference call, thereby reducing the frequency of heavy model inference \cite{OpenVLA}. To accommodate this in the ART framework, we execute  tool‑use reasoning at every \(H\) timesteps: rather than invoking tool selection at every single control step, the model predicts reasoning tokens \(\boldsymbol{a}_{\mathrm{r},t}\) and tool‑use decision \(\boldsymbol{a}_{\mathrm{t},t}\) once per chunk. Then the selected tools are activated for next \(H\) timestep observations \(\tilde{\boldsymbol{o}}_{t:t+H}\).

\begin{figure*}[htbp]
	\centering
	\includegraphics[width=0.85\textwidth]{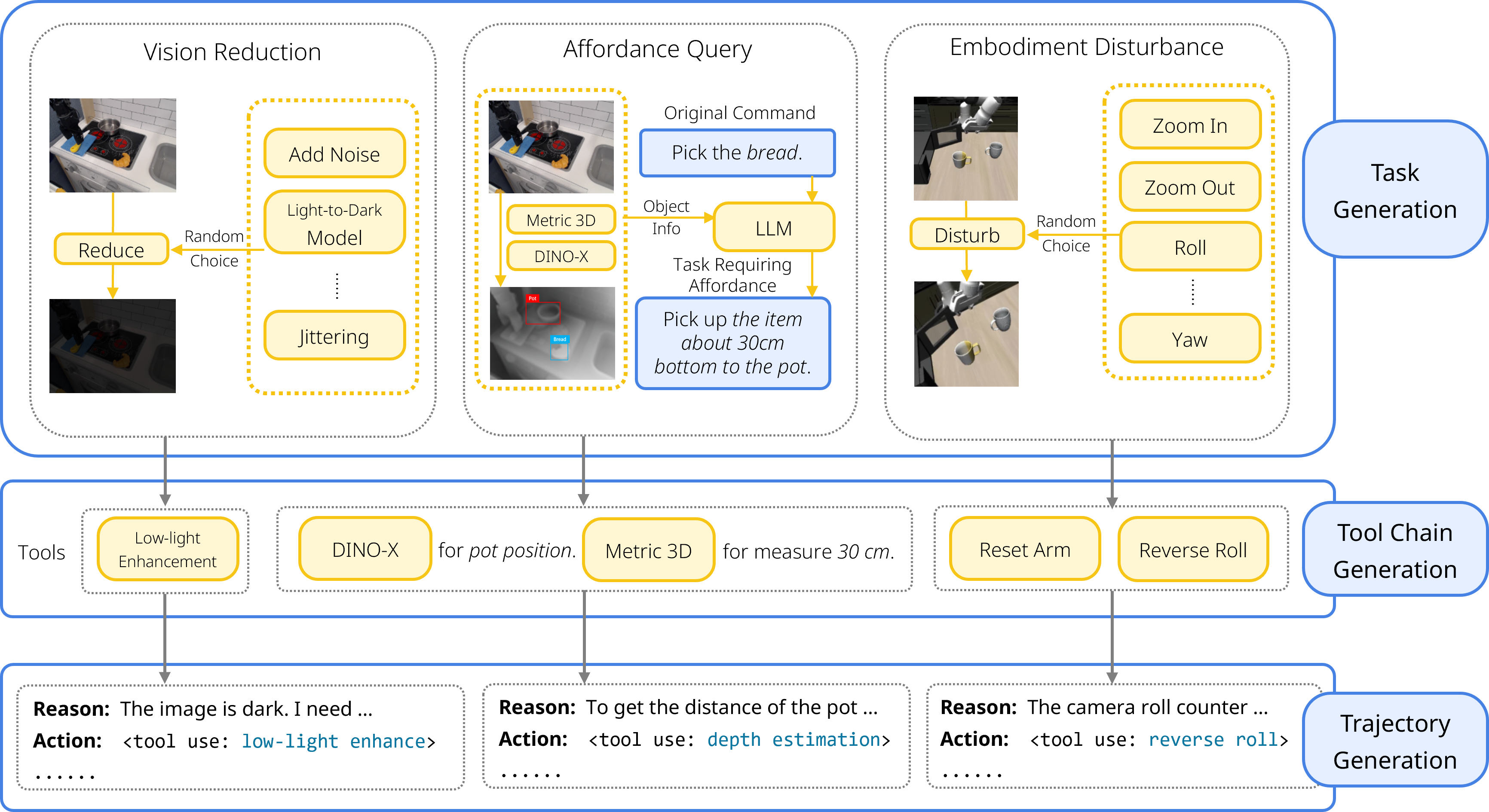}
	\caption{\textbf{Overview of \oursicon{} dataset collection pipeline.} The core idea is to fully utilize existing robotic dataset. (1) \textbf{Task generation}. For the vision and embodiment modalities, we randomly select some disturbances. For the language modality, we use GPT to transform simple tasks from the original dataset into reasoning tasks that require affordance information. (2) \textbf{Tool chain generation}. The tools needed to resolve disturbances and affordance reasoning are recorded into a reasoning chain. (3) \textbf{Trajectory generation}. A GPT is prompted to chain these models with the reasoning together.}
	\label{Fig: Dataset Collection}
\end{figure*}

\subsection{Training Procedures}\label{Sec: Training Procedures}

We use the pre-trained 3B \(\pi\)-FAST \cite{PI-FAST} model. During training, we optimize two components, the VLA backbone, and the finetuning LoRA module. As is shown in Sec.~\ref{Sec: Optimization with Tool-use Reasoning}, our training objective has two key parts: tool use reasoning token prediction and embodied action token prediction.

\paragraph{Tool use reasoning token prediction.}
For tool use reasoning, the model first predicts the discrete language tokens of reasoning \(\boldsymbol{a}_{\mathrm{r},t}\) and generates a discrete set of tool use actions \(\boldsymbol{a}_{\mathrm{t},t}\) based on the current raw observations \(\tilde{\boldsymbol{o}}_t\). This prediction is treated as a token selection problem and trained using next-token prediction loss, formulated as
\begin{equation}\label{Eq: Optimization Objective Decomposition}
	\mathcal{L}_\mathrm{tool} = - \sum_{t=1}^T \log P_{\theta_\mathrm{L}}(\boldsymbol{a}_{\mathrm{r},t}, \boldsymbol{a}_{\mathrm{t},t} | \tilde{\boldsymbol{o}}_{1:t}),
\end{equation}
where \(\theta_\mathrm{L}\) denotes parameters of the LoRA modules.

\paragraph{Embodied action token prediction.}
For embodied action prediction, the model generates a sequence of embodied actions \(\boldsymbol{a}_t\) in an auto-regressive manner, using the enhanced observations \(\boldsymbol{o}_t\). Since FAST \cite{PI-FAST} discretizes the continuous signal using discrete cosine transform, the transformed action tokens are trained with auto-regressive next-token prediction. The loss function for embodied actions is given by:
\begin{equation}
	\mathcal{L}_\mathrm{action} = - \sum_{t=1}^T \log P_{\theta}(\boldsymbol{a}_t | f_{t-1}(\tilde{\boldsymbol{o}_t})),
\end{equation}

where \(\theta\) represents the parameters of the VLA model without LoRA modules, and \(f_{t-1}\) is the activated function at the previous timestamp. This formulation allows for efficient training, ensuring that each action token is predicted based on the enhanced observations, resulting in improved robot behavior.


\section{Data Collection}\label{Sec: Data Collection}

Compared to the Internet-scale multi-modal datasets used in other multi-modal model research, a common challenge in VLA research is the high cost of collecting robotic datasets. Following the approach of T3-Agent, we propose an extension to the currently available VLA datasets, \oursdata{} (\textbf{A}ction with \textbf{T}ool), incorporating reasoning and tool use into the existing datasets. To the best of our knowledge, we are the first to introduce a VLA dataset with long-trajectory tool-use reasoning. Below, we will detail the definition of tools (\ref{Sec: Multi-modal Tool Enhancement}) and the process of long-trajectory generation (\ref{Sec: Long-Trajectory Tool Reasoning Generation}). Statistics and the examples of our dataset will be listed in Appendix.

\subsection{Multi-modal Tool Enhancement}\label{Sec: Multi-modal Tool Enhancement}

We define tools as modifications and enhancements to observations. As shown in Eq. \ref{Eq: Standard VLA}, the VLA model accepts three types of inputs: visual data \(\boldsymbol{I}_t\), language instructions \(\boldsymbol{l}\), and the proprioceptive states \(\boldsymbol{q}_t\) of the robot. We categorize the tools into visual, affordance, and embodiment enhancements, which correspond to the three input modalities: (1) \textbf{Visual Enhancement.} We define 10 types of visual enhancement tools, including low-light enhancement, denoising, jitter correction, and deblurring. These tools assist the model in adapting to more robust and complex scenarios. (2) \textbf{Affordance Enhancement.} We provide the \oursicon{} model with tools for depth estimation \cite{Metric3D} and object detection \cite{DINO-X}. These tools help the model effectively determine affordance information. For example, when handling the task of identifying the "farthest object," the model can use affordance enhancement to pinpoint the specific target. (3) \textbf{Embodiment Enhancement.}We consider a simple robotic setup, including the movement of the head-mounted camera and the initial state of the robotic arm. As pointed out by LIBERO-Plus \cite{LIBERO-Plus}, the performance of models significantly degrades when there is a shift in the viewpoint or initial state. We provide the robot with tools for camera rotation, zooming, and resetting the robot's bodily state, allowing the model to learn how to use these tools to solve problems.

\subsection{Long-Trajectory Tool Reasoning Generation}\label{Sec: Long-Trajectory Tool Reasoning Generation}

In this subsection, we detail the process of generating long-trajectory tool-use reasoning tasks. The task generation process begins by creating a series of challenges that require the model to reason about its environment and the appropriate tool to use. These challenges are designed using the three types of tool enhancements: visual, affordance, and embodiment, which correspond to different task types.

\paragraph{Task Generation.} 
In the task generation stage, the model is tasked with solving specific challenges that require tool use and reasoning. This includes three primary categories: \textbf{Vision Reduction}, where random visual degradations (such as noise or light transitions) are applied to simulate difficult environmental conditions, and the model uses visual enhancement tools to compensate; \textbf{Affordance Query}, where the model must identify objects or properties, with the help of tools like depth estimation and object detection; and \textbf{Embodiment Disturbance}, where disturbances in the robot's embodiment (e.g., shifts in viewpoint or initial state) are simulated, and the model uses tools like camera rotations or position resetting to address these changes.

\paragraph{Tool Chain Generation.}
Once a task is generated, it is paired with the corresponding tools necessary to solve the task. This involves creating a tool chain, a sequence of tasks and the tools required to complete them. For example, a task might require a vision enhancement tool followed by an embodiment adjustment tool, creating a specific tool chain that the model must follow.

\begin{figure*}[htbp]
	\centering
	\includegraphics[width=0.9\textwidth]{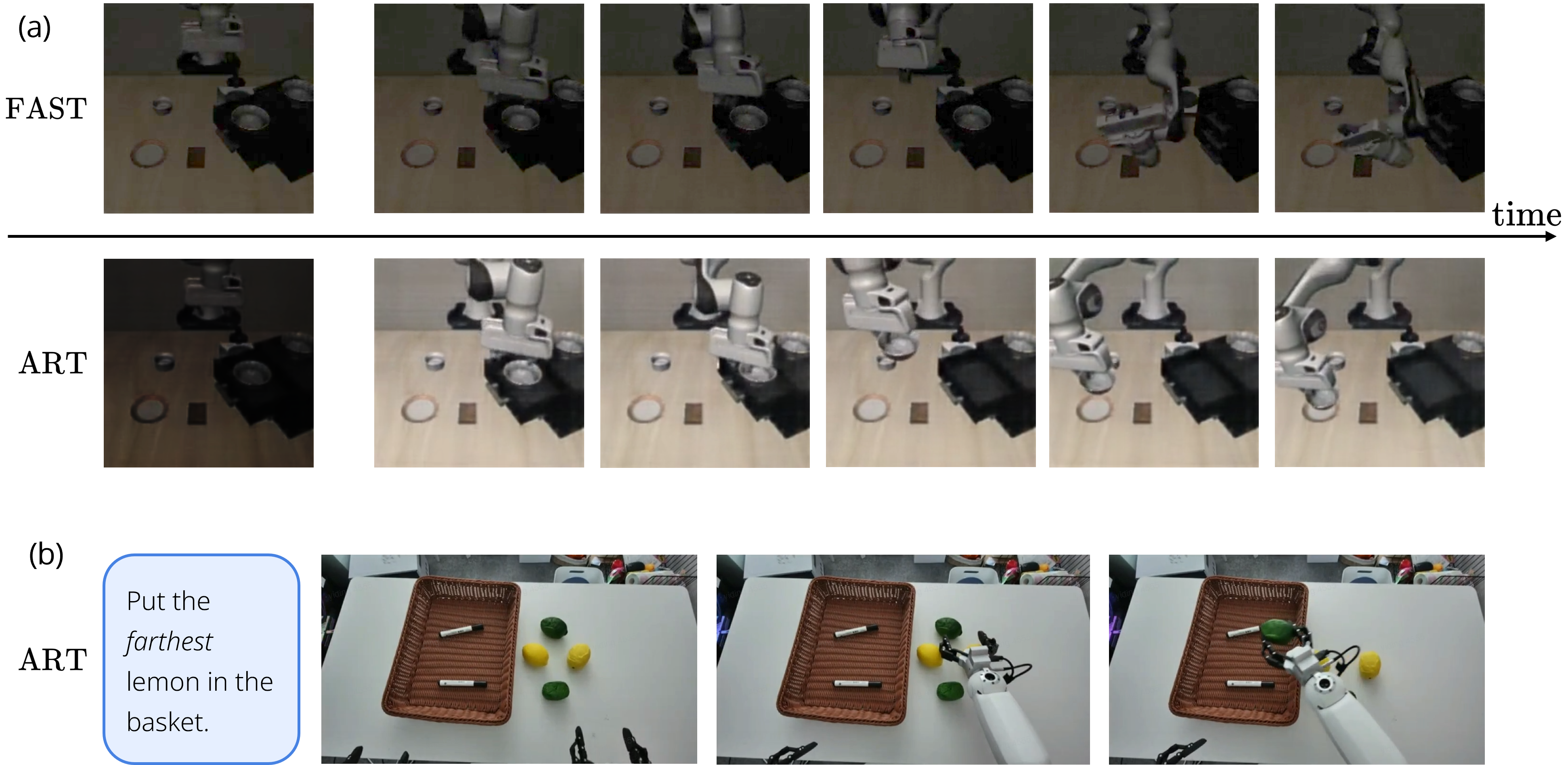}
	\caption{\textbf{Rollouts of experiments.} (a) The results of \(\boldsymbol{\pi}_0\)-FAST and \oursicon{} on low-light environment. (b) The performance of \oursicon{} on affordance reasoning task, whereas the robot controlled by \(\boldsymbol{\pi}_0\)-FAST predicts no action tokens and remains static here.}
	\label{Fig: Roolout}
\end{figure*}

\paragraph{Trajectory Generation.}
The final stage in the process is the generation of a long-trajectory reasoning process. Here, the task-tool pair is used to prompt a model (such as GPT) to generate a detailed reasoning trajectory. This trajectory explains why each tool is needed and how it helps complete the task. The model's reasoning is then consolidated into a coherent, step-by-step process that forms a long trajectory, guiding the robot through each action and decision point.

\section{Experiment}

The \oursicon{} model is designed to generalize broadly to new environments, tasks, and actions with extendable tools. In this section, we evaluate the effectiveness of \oursicon{} in utilizing external tools through tool-use reasoning. 


\begin{table*}[htbp]
	\centering
	\small
	\renewcommand{\arraystretch}{1.2}
	\setlength{\tabcolsep}{7pt}
	
	\begin{tabular}{c cccc cccc}
		\toprule
		\multirow{2}{*}{\textbf{Tasks}} & \multicolumn{4}{c}{\textbf{\oursdata{} LIBERO} (Success Rate) \(\uparrow\)} & \multicolumn{4}{c}{\textbf{\oursdata{} Astribot S1} (Success Rate) \(\uparrow\)} \\
		\cmidrule(lr){2-5} \cmidrule(lr){6-9}
		& Vision & Affordance & Embodiment & \textbf{Average} &
		Vision & Affordance & Embodiment & \textbf{Average} \\
		\midrule
		\textbf{OpenVLA} &
		20\% & 10\% & 7\% & 12\% & 5\% & 5\% & 10\% & 6.7\% \\
		\midrule
		\textbf{\(\boldsymbol{\pi}_0\)} &
		65\% & 15\% & 10\% & 30\% & 40\% & 30\% & 40\% & 37\% \\
		\midrule
		\textbf{\(\boldsymbol{\pi}_0\)-FAST} &
		60\% & 12\% & 45\% & 39\% & 40\% & 30\% & 60\% & 43\% \\
		\midrule
		\textbf{\oursicon{}-FAST} &
		\textbf{81\%} & \textbf{62\%} & \textbf{82\%} & \textbf{75\%} & \textbf{70\%} & \textbf{55\%} & \textbf{70\%} & \textbf{62\%} \\
		\midrule
	\end{tabular}
	\caption{
		\textbf{Performance improvement by on-the-fly tool use.} We compare the success rate of generalized tasks between \oursicon{} and different mainstream VLA models on different generalized tasks. The LIBERO experiments show that tool use significantly improve the model's ability in complex scenarios. The Astribot S1 experiments show the tool reasoning ability tuned on \oursdata{} can generalize to new robotic scene.
	}
	\label{tab:main_results}
\end{table*}

\subsection{Experiment Setup}

In this section, we introduce our training settings and describe the three benchmark environments used for model evaluation and comparison. These include a simulated environment, a closed-loop real-world scenario, and an open-loop real-world testing setup.


\vspace{-1em}

\paragraph{Training Settings.} Following the previously designed Training Injection method, we fine-tuned the model for 1 epoch on a 30k tool trajectory dataset introduced in Sec.~\ref{Sec: Data Collection}, starting from a pre-trained model. The fine-tuning was conducted on \(8 \times\) A800 GPUs with a batch size of 24. An initial learning rate of \(5 \times 10^{-5}\) was used, with a 1k-step warm-up period.

\vspace{-1em}

\paragraph{Astribot S1.} We evaluated our model on the Astribot S1 humanoid robot, a dual-arm robot with 16 degrees of freedom. Initially, we trained a baseline FAST model on a 16k pick-and-place action trajectory dataset, which includes 80 different objects and 10 container types, following standard VLA training formats. 

\vspace{-1em}

\paragraph{\oursdata{} Dataset (LIBERO).} We tested our model in the LIBERO environment, a simulated platform that allows for modifications in visual factors such as lighting and noise. The precision of task instructions was also adjusted based on object positioning (e.g., changing "place on the plate" to "place 30 cm to the left of the drawer"). In addition, disturbances were introduced to the initial positions of the camera and the robot's body. These modifications provided a controlled environment for comparing the performance of \oursicon{} with current SOTA baselines in simulation.

\subsection{Performance of Tool Integration}

Table~\ref{tab:main_results} shows the straightforward improvement brought by \oursdata{}. As reported by LIBERO-Plus~\cite{LIBERO-Plus}, state-of-the-art (SOTA) Vision-Language-Action (VLA) models struggle significantly with visual, semantic, and embodiment disturbances. These disturbances reduce their performance on complex tasks, especially when compared to general robotics datasets that contain high-quality, clean observations. In our experiments, models like OpenVLA and \(\boldsymbol{\pi}_0\) exhibit poor performance, with average success rates significantly below 30\% across both the LIBERO simulated environment and real-world Astribot S1 robot tasks. This demonstrates the difficulty of applying these models to real-world scenarios, where noise and variability in observations can dramatically affect the results.

\vspace{-1em}

\paragraph{Performance of on-the-fly tool use.} The key advantage of \oursicon{} is its ability to integrate external tools on-the-fly, enabling it to adapt to a variety of complex tasks without the need for retraining the entire model. This tool-based reasoning capability allows \oursicon{} to address the disturbances that hinder the performance of traditional VLA models. As seen in Table~\ref{tab:main_results}, \oursicon{}-FAST outperforms other models in both the LIBERO environment and the Astribot S1 robot tasks, with an average success rate of 75\% and 62\%, respectively. This significant improvement suggests that external tools can be leveraged effectively for real-time decision-making, allowing \oursicon{} to handle a wider range of tasks, even in environments with noisy or incomplete observations. This capability demonstrates the potential of on-the-fly tool use in enhancing the generalization of VLA models to more complex and dynamic scenarios.

\vspace{-1em}

\paragraph{Generalization of tool reasoning ability.} Another notable result is the ability of \oursicon{} to generalize tool-use reasoning across multiple environments. While previous SOTA models struggle to generalize beyond their training domains, \oursicon{} shows robust performance in both the LIBERO simulation and real-world robotic tasks. The fine-tuning on the \oursdata{} dataset, which includes data from LIBERO \cite{LIBERO}, Bridge v2 \cite{BridgeData-v2}, and DROID \cite{DROID}, enables the model to effectively apply tools across different datasets, achieving impressive results in new tasks without requiring task-specific re-training. This confirms that the tool reasoning ability of \oursicon{} is not only effective in controlled environments but also transferable to real-world applications, where the robot must adapt to new objects, goals, and conditions. The ability to generalize this tool-use capability is a crucial step towards building more adaptable and robust robotic systems.

\subsection{Performance of Offloading Model Tasks}

\begin{table}[htbp]
	\centering
	\small
	\renewcommand{\arraystretch}{1.2}
	
	\begin{tabular}{c ccc}
		\toprule
		\multirow{3}{*}{\makecell{\textbf{Affordance} \\ \textbf{Tasks}}} & \multicolumn{3}{c}{\textbf{\oursdata{} LIBERO} (Success Rate) \(\uparrow\)} \\
		\cmidrule(lr){2-4}
		& \makecell{w/ Vision \\ Corruption} & \makecell{w/o Vision \\ Corruption} & \textbf{Average} \\
		\midrule
		\textbf{ECoT} &
		12\% & 58\% & 36\% \\
		\midrule
		\textbf{\oursicon{}-FAST} &
		\textbf{72\%} & \textbf{62\%} & \textbf{67\%} \\
		\midrule
	\end{tabular}
	
	\caption{
		\textbf{Performance improvement compared to the reason-only model.} We test the success rate of ECoT and \oursicon{} on the affordance part of \oursdata{}, with or without vision corruptions.
	}
	\label{tab:ECoT}
\end{table}

\paragraph{Comparison with End-to-End VLA Models.} In this section, we compare the tool-use approach with traditional end-to-end VLA models. Specifically, we consider two types of baseline models: Action-only models (Fig.~\ref{Fig: Pipeline Comparison.} (1a)) and Reason-only models (Fig.~\ref{Fig: Pipeline Comparison.} (1b) \& (2b)).

\begin{table}[htbp]
	\centering
	\small
	\renewcommand{\arraystretch}{1.2}
	
	\begin{tabular}{c ccc}
		\toprule
		\multirow{2}{*}{\textbf{Tasks}} & \multicolumn{3}{c}{\textbf{\oursdata{} LIBERO} (Success Rate) \(\uparrow\)} \\
		\cmidrule(lr){2-4}
		& Vision & Embodiment & Affordance \\
		\midrule
		\textbf{FAST} (Post-trained) &
		71\% & 61\% & 65\% \\
		\midrule
		\textbf{\oursicon{}-FAST} &
		\textbf{81\%} & \textbf{62\%} & \textbf{82\%} \\
		\midrule
	\end{tabular}
	
	\caption{
		\textbf{Performance improvement compared to end-to-end training.} We test the success rate of \oursicon{} and the finetuned FAST model with the same training consumption.
	}
	\label{tab:End-to-end}
\end{table}

\vspace{-1em}

\paragraph{Comparison with end-to-end learning.} A standard VLA model typically relies on an end-to-end learning paradigm. This approach assumes that, given enough data, the model can learn the input-output mapping directly from raw observations to actions through a data-driven process. To evaluate the effectiveness of \oursicon{} in comparison to such models, we fine-tuned a FAST model with the same amount of training data, but removed the intermediate tool-invocation reasoning. In this case, the model was tasked with learning to generate embodied actions directly from raw observations. While the FAST model exhibited some improvement, it still lagged behind \oursicon{} in terms of its ability to handle tasks using external tools. As shown in Table~\ref{tab:End-to-end}, \oursicon{} significantly outperforms the post-trained FAST model, demonstrating that incorporating external tools allows for more robust task-solving and better generalization to unseen scenarios.

\vspace{-1em}

\paragraph{Comparison with ECoT.} We also compared \oursicon{} with another common reasoning-only model in VLA research, ECoT~\cite{ECoT}.As shown in Table~\ref{tab:ECoT}, \oursicon{} performed slightly better than ECoT when no vision corruption was introduced, validating that \oursicon{} can handle affordance tasks more effectively when the input data is clean. ficant finding emerges when vision disturbances are added. ECoT's performance significantly drops, as it was only trained on high-quality, clear data without exposure to such visual distortions. This causes the model to encounter out-of-distribution data during inference, severely limiting its generalization capability. In contrast, \oursicon{} maintains robust performance by leveraging external vision enhancement tools, which help to mitigate the effect of these disturbances. This approach improves task generalization without requiring retraining, thus offering a flexible solution to complex, dynamic environments.

\section{Conclusion}

This paper presents ART (Agentic Robot with Tool-use), a fine-tuning framework that bridges modular and end-to-end VLA models. It enables VLA models to leverage external tools for novel tasks without compromising their precision or continuous action capabilities. Our method integrates multi-modal tool usage into existing VLA datasets, reducing the need for new data collection, and employs a two-stage LoRA fine-tuning process to decouple tool reasoning from core VLA tasks, mitigating catastrophic forgetting. Experimental results in both the LIBERO simulation and real-world environments show that ART enhances model robustness and generalizability, enabling effective handling of new objects, scenarios, and action instructions. ART reduces data dependency, paving the way for more adaptable robotic agents in dynamic environments.

{
	\small
	\bibliographystyle{ieeenat_fullname}
	\bibliography{main}
}

\end{document}